\documentclass{article}
\usepackage[preprint]{neurips_2026}
\workshoptitle{The 9th Workshop on Machine Learning and the Physical Sciences (ML4PS 2026)}
\makeatletter
\renewcommand{\@noticestring}{Author draft for ML4PS 2026. Do not distribute.}
\makeatother
\usepackage[utf8]{inputenc}
\usepackage[T1]{fontenc}
\usepackage{amsmath,amssymb,booktabs,microtype,graphicx}
\usepackage{hyperref}
\hypersetup{hidelinks,pdfauthor={Rashmita Soni, Abhirup Ghosh, Soumya Banerjee},pdftitle={Query-ATLAS: Symbolic Discovery of Physical Laws across Distributed Observations}}

\title{Query-ATLAS: Symbolic Discovery of Physical Laws across Distributed Observations}
\author{%
Rashmita Soni$^{1}$ \quad Abhirup Ghosh$^{2}$ \quad Soumya Banerjee$^{3}$\\
$^{1}$Indian Institute of Technology Delhi, India\\
$^{2}$University of Birmingham, UK\\
$^{3}$University of Cambridge, UK\\
\texttt{rashmita.soni@alumni.iitd.ac.in}\\
\texttt{a.ghosh.1@bham.ac.uk} \quad \texttt{neel.soumya@gmail.com}
}
\begin{document}
\maketitle

\begin{abstract}
Measurements of a shared physical law may be distributed across experiments whose
raw observations cannot be pooled. We study whether adaptive coordination of
symbolic search and language-model hypotheses helps recover one global equation
from such evidence. Query-ATLAS selects tools for profiling, structural proposal,
local symbolic regression, and cross-client scoring through an equation-level
interface. Across 12 Feynman benchmarks, two client distributions, and three seeds,
we compare five architectures in 360 runs. With overlapping support, Query-ATLAS
recovers 23 of 36 equations exactly, versus 21 for local PySR with cross-client
selection; both additional recoveries concern relativistic mass. With partitioned
support it recovers 18 versus 19, and its mean runtime is about five times higher.
These results identify a limited benefit from scientific priors, while exposing
the gap between accurate approximation and equation recovery. They motivate
controlled evaluation of adaptive search rather than a claim of general superiority.
\end{abstract}

\section{Introduction}
Symbolic regression seeks an analytic expression explaining observations and is
a useful tool for scientific model discovery \citep{cranmer2023}. In a distributed
setting, laboratories or instruments may measure the same variables over different
operating ranges while retaining their raw records. An expression accurate within
one range may be an approximation to a shared law rather than the law itself.

Neural-guided population seeding combines learned proposals with genetic
programming \citep{mundhenk2021}. LLM-SR combines scientific priors, equation
programs, coefficient fitting, and evolutionary search \citep{shojaee2025sr}.
In-Context Symbolic Regression refines functional forms using LLM feedback and
an external coefficient optimizer \citep{merler2024}; LLM4ED alternates
LLM-guided self-improvement and evolutionary search for dynamical equations
\citep{du2024}. SR-Scientist further studies autonomous tool use for equation
discovery \citep{xia2026}. These works motivate combining structural proposals
with data-based evaluation; we do not claim that combination as new.

Distributed symbolic regression is also established: Bayesian Federated Symbolic
Regression combines model selection with distributed parameter inference
\citep{billa2025}. Our focus is adaptive tool selection across distributed
observations through an interface exchanging equations and aggregate scores
rather than raw observations.

\paragraph{Problem statement.}
Given clients observing different ranges of the same unknown physical law, recover
one compact equation supported across clients. Under a fixed computational budget,
does evidence-dependent tool selection improve recovery and generalization over
repeated local search and a fixed schedule with the same tools? The implemented controller and paired analysis across two support regimes provide
an exploratory comparison; unequal computation and evidence access prevent causal
attribution to adaptive orchestration alone.

\begin{figure}[t]
\centering
\includegraphics[width=\linewidth]{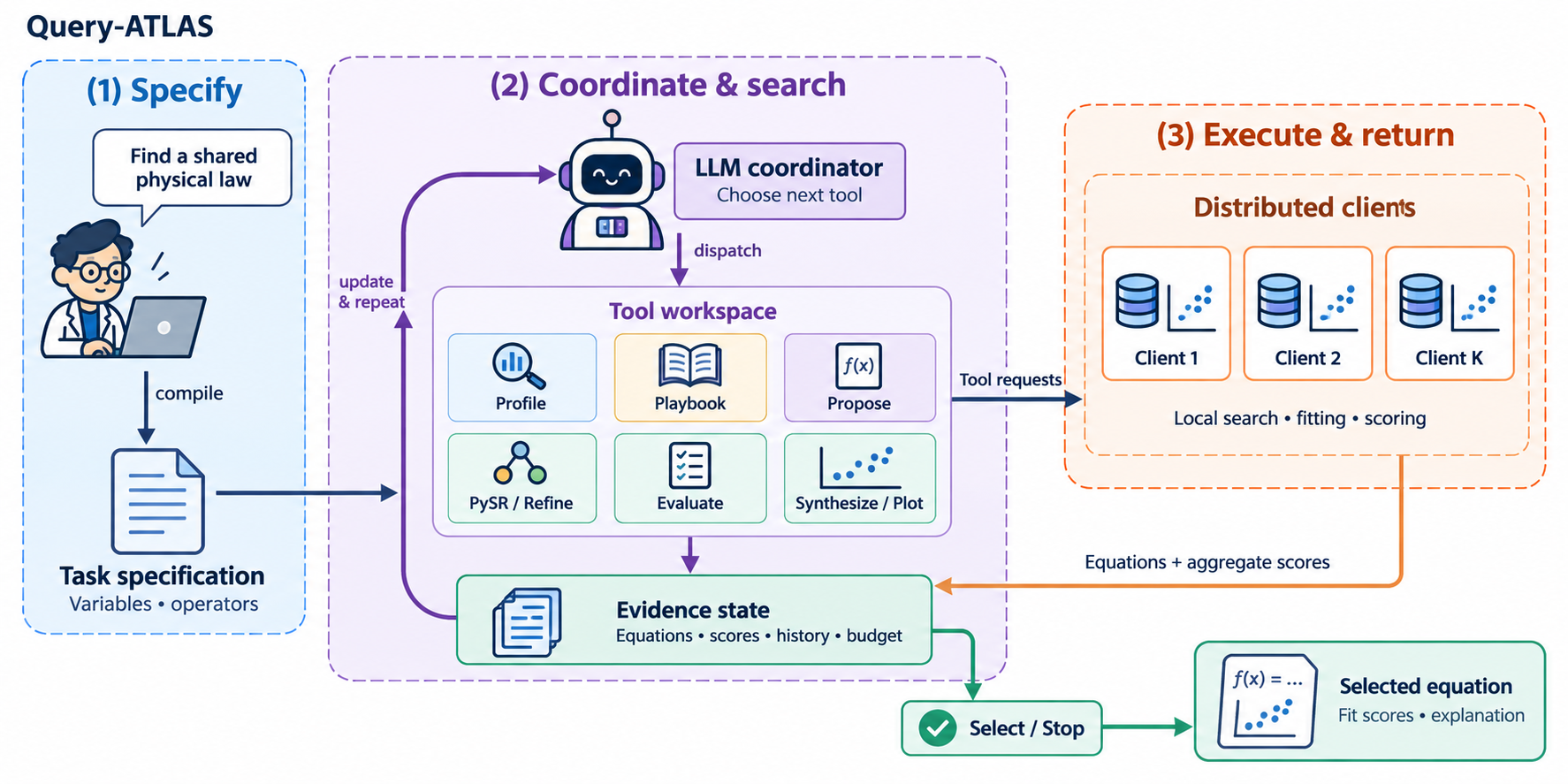}
\caption{Query-ATLAS workflow: compile a task, select tools, and collect client
evidence to repeat or stop. The robot denotes one LLM coordinator; icons denote
tools and client workers. Arrows show the discovery interface. Final fitting
currently pools client rows in memory (Section~2).}
\label{fig:architecture}
\end{figure}

\section{Method and information boundary}
Client $k$ holds $D_k=\{(\mathbf{x}_{ki},y_{ki})\}_{i=1}^{n_k}$ from a shared
unknown equation $f^\star$, with potentially different feature support across
clients. The discovery interface returns feature ranges, counts, moments,
candidate expressions, and aggregate fit scores. Let
$e_k(f)=\mathrm{RMSE}_{D_k}(f)/\mathrm{std}_{D_k}(y)$ and
$\bar e(f)=K^{-1}\sum_k e_k(f)$. For effectively constant targets, the
implementation uses unnormalized RMSE. A candidate is numerically valid if it
produces finite predictions on every observed client row.

\paragraph{Adaptive controller.}
Figure~\ref{fig:architecture} shows the intended feedback loop. The controller
observes the compiled variable specification, equation catalog,
action history, scores, and remaining budgets. At each step it selects a permitted
tool and records a brief decision summary. \textsc{Profile} returns client
statistics. \textsc{Playbook} retrieves compatible structural families and domain
cautions. \textsc{Propose} produces LLM equation hypotheses, including parameterized
skeletons. \textsc{Refine} fits their coefficients locally using BFGS and searches
for additive residual corrections with PySR. \textsc{PySR} independently searches
each client's observations. \textsc{Evaluate} scores candidates across clients.
\textsc{Synthesize} samples local surrogate equations over reported ranges to
form a diagnostic synthetic dataset and additional candidates. The controller may
nominate a candidate or stop; deterministic checks govern acceptance.

\paragraph{Final selection.}
For the all-client-valid candidate set $\mathcal V$, let
$e_{\min}=\min_{f\in\mathcal V}\bar e(f)$. The default selector is
\begin{equation}
\hat f=\underset{f\in\mathcal V:\,\bar e(f)\leq1.05e_{\min}}{\operatorname{lexmin}}
\left(C(f),\ \max_k e_k(f),\ \bar e(f)\right),
\label{eq:selection}
\end{equation}
where $C$ counts nodes in the simplified SymPy expression tree and
$\operatorname{lexmin}$ orders the three criteria from left to right. This describes
the implemented selection rule rather than a weighted surrogate objective.
The controller's nominated result must also pass numerical-validity and
near-best-error checks. The current scores reuse each client's fitting rows;
they are training-fit feedback, not independent local validation estimates.
Only the separate global test fold measures held-out prediction accuracy.

\paragraph{Scope of the boundary.}
The client API and model prompts exchange summaries and equations. This is a
simulated distributed interface, not a secure deployment or differential-privacy
guarantee. The final-fitting helper concatenates client rows in memory, so the
implementation does not enforce end-to-end data locality. Statistics and equations
may leak information. The CLI model adapter
uses a read-only sandbox; this does not establish filesystem isolation from the
benchmark repository. Numerical validity on observed rows also does not establish
dimensional consistency or validity throughout a physical domain.

\section{Experimental protocol and comparison metrics}
We use 12 Feynman datasets available through SRBench/PMLB \citep{lacava2021}:
I.6.2a, I.8.14, I.10.7, I.11.19, I.12.1, I.13.4, I.18.4, I.27.6, I.34.14,
I.40.1, I.44.4, and II.11.27. They include Gaussian, geometric, relativistic,
mechanical, exponential, logarithmic, and rational relationships. Each dataset
contains 100,000 rows. We split it into 75,000 training and 25,000 test rows before
dividing training data among five clients (15,000 rows each). In the overlapping
setting, disjoint IID shards have overlapping feature support. In the partitioned
setting, training rows are sorted by the first feature and divided into contiguous
shards. The global random test fold is the same for both settings at a given seed;
it is not an extrapolation test outside the training population. There is no added
noise. Seeds 11, 22, and 33 give 36 tasks per setting and method.

\paragraph{Architectures and resources.}
All methods share each task's operator vocabulary and PySR settings: 150 iterations,
eight populations, maximum expression size 40, and serial deterministic search
within each process. Independent client processes use two workers. Query-ATLAS has
caps of 20 LLM calls, 100 searches, 35 agent steps, and five proposal rounds;
each refinement action handles at most two skeletons. Local PySR performs one
search per client, pools the candidate expressions, and applies cross-client
selection. Centralized PySR performs one search on pooled training rows and returns
the candidate with lowest macro client error; it is a pooled-data reference, not
an upper bound. Fixed LLM feedback performs four rounds of proposal, local
refinement, and scoring, with three proposals per round. Range-synthetic PySR
performs local searches and fits another search to synthetic surrogate outputs.
Its current implementation takes the first returned local candidate, making its
results sensitive to candidate ordering. The fixed loop also has a different
profile/playbook interface from Query-ATLAS, so it is not a matched tool-policy
ablation. LLM calls use GPT-5.6 Sol through the Codex CLI, as confirmed by the
authors. The LLM is used for inference without fine-tuning. The frozen run metadata
does not record the exact model snapshot, limiting exact reproduction.

\paragraph{Primary and supporting metrics.}
Our primary endpoint is \emph{exact symbolic recovery}: the stored evaluator
counts a success when SymPy simplifies $\hat f-f^\star$ to zero. We report counts
over 36 tasks and paired outcomes on identical equation--seed instances.
This strict test can miss equivalent expressions with rounded coefficients or
identities requiring additional domain assumptions. We separately report test
NRMSE, expression complexity, runtime, and actual tool use. Test NRMSE divides
test RMSE by the standard deviation of test targets. The earlier relaxed scorer
can accept additive offsets or multiplicative rescaling and uses numerical checks
for some nonlinear cases; we therefore exclude it from the main recovery endpoint.
We do not combine recovery and cost into an arbitrarily weighted score, nor claim
these runs compare recovery at a matched computational budget.

\begin{table}[t]
\caption{Completed results; 36 tasks per setting and architecture. Exact is the
number of symbolic recoveries. NRMSE, complexity ($C$), and seconds are arithmetic
means. Computation is not matched across architectures.}
\label{tab:main}
\centering
\begin{tabular}{lrrrr}
\toprule
Method & Exact & Test NRMSE & $C$ & Seconds\\
\midrule
\multicolumn{5}{l}{\textit{Overlapping client support}}\\
Query-ATLAS & 23/36 & 0.001284 & 16.97 & 241.8\\
Local PySR & 21/36 & 0.001287 & 17.64 & 51.5\\
Centralized PySR & 15/36 & 0.030643 & 26.53 & 30.6\\
Fixed LLM feedback & 7/36 & 0.234096 & 17.42 & 609.2\\
Range-synthetic PySR & 0/36 & 0.858940 & 1.94 & 62.5\\
\midrule
\multicolumn{5}{l}{\textit{Partitioned client support}}\\
Query-ATLAS & 18/36 & 0.079127 & 16.78 & 221.3\\
Local PySR & 19/36 & 0.039859 & 17.72 & 43.1\\
Centralized PySR & 14/36 & 0.029358 & 26.22 & 25.1\\
Fixed LLM feedback & 4/36 & 0.245961 & 17.64 & 526.5\\
Range-synthetic PySR & 0/36 & 0.662625 & 6.22 & 55.3\\
\bottomrule
\end{tabular}
\end{table}

\section{Results: a narrow recovery benefit at higher cost}
Table~\ref{tab:main} shows higher overlapping-support recovery for Query-ATLAS
(63.9\%) than local PySR (58.3\%), with nearly identical mean test error.
The partitioned setting reverses the recovery ordering (50.0\% versus 52.8\%)
and gives Query-ATLAS approximately twice the mean test NRMSE.
These are descriptive results on 12 equations with three seeds each; 360 total
architecture runs should not be interpreted as 360 independent scientific problems.

\paragraph{Where the difference comes from.}
Paired comparison localizes both overlapping-support gains to relativistic mass,
$m=m_0/\sqrt{1-v^2/c^2}$, at seeds 11 and 22. In these runs, the selected
Query-ATLAS expressions have LLM provenance and test NRMSE of approximately
$4.3\times10^{-16}$. Local PySR instead selects longer rational approximations
with NRMSE $4.27\times10^{-5}$ and $6.19\times10^{-5}$: accurate predictions
without exact law recovery. At seed 33 both methods select the same non-exact
approximation. All other overlapping equation--seed recovery outcomes agree.
Under partitioning, Query-ATLAS has no additional exact recoveries and loses one
Euclidean-distance recovery at seed 22. This supports investigating scientific priors
as a way to distinguish a law from an approximation, but cannot distinguish useful
prior knowledge from recognition of a familiar benchmark equation.

\paragraph{Cost and interpretation.}
Query-ATLAS takes 4.70 times the local baseline's mean runtime with overlapping
support and 5.14 times with partitioning. Its mean search counts are 13.33 and
13.78, versus five for local and one for centralized PySR; mean LLM call counts
are 8.92 and 8.22. Fixed feedback uses four LLM calls and 60 searches per task.
Search-call counts alone are not equivalent computational budgets because data
sizes and residual problems differ. The pooled and synthetic references also have
different candidate-selection behavior. Their weaker recovery here should not be
read as evidence that centralization or synthetic aggregation is inherently inferior.

\section{Limitations and conclusion}
The study covers noiseless, familiar equations with many observations per client;
it establishes neither sparse-data performance nor discovery of previously unknown
physics. Physical variable names and task-specific operators supply useful priors
to the search. A planned extension uses a prespecified physics/materials subset of LSR-Synth
from LLM-SRBench \citep{shojaee2025bench,xia2026}, preserving its training, in-domain,
and out-of-domain test splits and partitioning only training observations.
Alongside symbolic recovery and test NRMSE, we propose accuracy-to-tolerance at
1\% and 0.1\%, following SR-Scientist's removal of the worst 5\% of relative errors,
with its zero-target handling verified before comparison. Worst-client held-out
NRMSE and recovery versus runtime would test cross-client robustness and cost.
These proposed evaluations require model isolation from benchmark answers,
independent client validation, and matched computation and tools.

The completed comparison finds two additional recoveries of one law under
overlapping support, offset by higher cost and a failure under partitioning.
The research direction is selective use of scientific priors with explicit
evidence and cost accounting.

\paragraph{AI assistance.}
LLMs generate discovery hypotheses and assisted manuscript editing, result auditing,
and figure scripting; an image-generation model illustrated Figure~\ref{fig:architecture}.
We checked its labels and arrows against the method, and checked saved results,
implementation, and references;
this revision is not an independent experimental rerun.

\end{document}